\documentclass[11pt,a4paper,dvipsnames]{article}
\usepackage[hyperref]{emnlp2020}
\usepackage{times}
\usepackage{latexsym}

\usepackage{graphicx}
\usepackage{colortbl}
\usepackage{multirow}
\usepackage{dingbat}
\usepackage{pifont}
\newcommand{\cmark}{\ding{51}}%
\newcommand{\xmark}{\ding{55}}

\definecolor{lgreen}{RGB}{73,174,137}
\definecolor{lred}{RGB}{182,49,54}
\definecolor{lorange}{RGB}{255, 128, 0}

\colorlet{lorange1}{lorange!20}

\colorlet{lgreen4}{lgreen!100}
\colorlet{lgreen3}{lgreen!60}
\colorlet{lgreen2}{lgreen!40}
\colorlet{lgreen1}{lgreen!20}
\colorlet{lgreen0}{lgreen!10}

\colorlet{lred4}{lred!75}
\colorlet{lred3}{lred!60}
\colorlet{lred2}{lred!40}
\colorlet{lred1}{lred!20}
\colorlet{lred0}{lred!5}

\usepackage{microtype}

\aclfinalcopy 

\title{BLEU might be Guilty but References are not Innocent}

\author{Markus Freitag, David Grangier, Isaac Caswell\\
  Google Research \\
  {\tt \{freitag,grangier,icaswell\}@google.com}}

\date{}

\begin{document}
\maketitle
\begin{abstract}
The quality of automatic metrics for machine translation has been
increasingly called into question, especially for high-quality systems.
This paper demonstrates that, while choice of metric is important, the nature
of the references is also critical. 
We study different methods to collect 
references and compare their value in automated evaluation by 
reporting correlation with human evaluation for a variety of systems
and metrics.
Motivated by the finding that typical references exhibit poor diversity,
concentrating around \textit{translationese} language, we develop a paraphrasing task for linguists to perform on existing reference translations, which counteracts this bias.
Our method yields higher correlation with human
judgment not only for the submissions of WMT 2019 English$\to$German, but also for 
Back-translation and APE augmented MT output, which have been shown to have low correlation with automatic metrics using standard references. We demonstrate that our methodology improves correlation with all modern evaluation metrics we look at, including embedding-based methods.
To complete this picture, we reveal that multi-reference BLEU does not improve the correlation for high quality output, and present an alternative multi-reference formulation that is more effective.
\end{abstract}

\section{Introduction}

Machine Translation (MT) quality has greatly improved in recent years
\cite{bahdanau2014attention,gehring2017conv,vaswani2017attention}.
This progress has cast doubt on the reliability of automated metrics, especially in the high accuracy regime.
For instance, the WMT English$\to$German evaluation in the last two years had
a different top system when looking at automated or human evaluation~\cite{bojar2018wmt,barrault2019wmt}.
Such discrepancies have also been observed in the past, especially when comparing rule-based 
and statistical systems \cite{bojar2016wmt_tenyear,koehn2006bleu,callison-burch2006bleu}.

Automated evaluations are however of crucial importance, especially for system 
development. Most decisions for architecture selection, hyper-parameter search 
and data filtering rely on automated evaluation at a pace and scale that would 
not be sustainable with human evaluations. Automated evaluation \citep{koehn2010book,papineni2002bleu} 
typically relies on two crucial ingredients: a metric and a reference translation. Metrics 
generally measure the quality of a translation by assessing the overlap between the system 
output and the reference translation. Different overlap metrics have been 
proposed, aiming to improve correlation between human and automated evaluations. Such 
metrics range from n-gram matching, e.g.\  BLEU~\citep{papineni2002bleu}, to accounting for synonyms, 
e.g.\ METEOR~\citep{banerjee2005meteor}, to considering distributed word representation, 
e.g.\ BERTScore~\citep{zhang19bertscore}. Orthogonal to metric quality~\citep{ma-etal-2019-results}, 
reference quality is also essential in improving correlation between human and automated 
evaluation.

This work studies how different reference collection methods impact the reliability of 
automatic evaluation.
It also highlights that the reference sentences typically collected 
with current (human) translation methodology are biased to assign higher automatic scores to MT output that share a similar style as the reference.
Human translators tend to generate
translation which exhibit \emph{translationese} language, i.e. sentences with source artifacts~\citep{koppel2011translationese}.
This is problematic because collecting only a single style of references fails to reward systems that might produce alternative but equally accurate translations~\cite{popovic2019reducing}. Because of this lack of diversity, multi-reference evaluations like multi-reference BLEU are also biased to prefer that specific style of translation. 

As a better solution, we show that paraphrasing translations, when done carefully, can improve the quality of automated evaluations more broadly. Paraphrased translations increase diversity and steer evaluation away from rewarding translation artifacts. Experiments with the official submissions of WMT 2019 English$\to$German for a variety of different metrics demonstrate the increased correlation with human judgement. Further, we run additional experiments for MT systems that are known to have low correlation with automatic metrics calculated with standard references. In particular, we investigated MT systems augmented with either back-translation or automatic post-editing (APE). We show that paraphrased references overcome the problems of automatic metrics and generate the same order as human ratings.

Our contributions are four-fold:
(i) We collect different types of references on the same test set and show that it is 
possible to report strong correlation between automated evaluation with human metrics, 
even for high accuracy systems.
(ii) We gather more natural and diverse valid translations by collecting human paraphrases of reference translations. 
We show that (human) paraphrases correlate well with human judgments when used as 
reference in automatic evaluations.
(iii) We present an alternative multi-reference formulation that is more effective than multi reference BLEU for high quality output.
(iv) We release\footnote{\url{https://github.com/google/wmt19-paraphrased-references}}
a rich set of diverse references to 
encourage research in systems producing other types of translations, and reward a wider 
range of generated language.

\section{Related Work}

Evaluation of machine translation is of crucial importance for system development and deployment decisions~\citep{moorkens2018book}. Human evaluation typically reports adequacy of translations, often complemented with fluency scores~\citep{white94arpaeval,graham-etal-2013-continuous}. 
Evaluation by human raters can be conducted through system 
comparisons, {\it rankings}~\citep{bojar2016wmt}, or absolute judgments, 
{\it direct assessments}~\citep{graham-etal-2013-continuous}. Absolute judgments 
allow one to efficiently compare a large number of systems. 
The evaluation of translations as isolated sentences, full paragraphs or documents is
also an important factor in the cost/quality trade-offs~\citep{carpuat2012trouble}. Isolated 
sentence evaluation is generally more efficient but fails to penalize contextual 
mistakes~\citep{tu2018doclevel,hardmeier-etal-2015-pronoun}.

Automatic evaluation typically collects human reference translations and relies on 
an {\it automatic metric} to compare human references to system outputs. Automatic
metrics typically measure the overlap between references and system outputs. A wide
variety of metrics has been proposed, and automated metrics is still an active area of 
research. BLEU~\citep{papineni2002bleu} is the most common metric. It measures the
geometric average of the precision over hypothesis n-grams with an additional penalty to
discourage short translations. NIST~\citep{doddington2002nist} is similar but considers 
up-weighting rare, informative n-grams. TER~\citep{snover06ter} measures an edit distance, 
as a way to estimate the amount of work to post-edit the hypothesis into the reference.
METEOR~\citep{banerjee2005meteor} suggested rewarding n-gram beyond exact matches, considering 
 synonyms. Others are proposing to use contextualized word 
embeddings, like BERTscore~\citep{zhang19bertscore}. 
Rewarding multiple alternative formulations is also the primary motivation behind
multiple-reference based evaluation~\citep{niessen2000evaluation}. 
\newcite{dreyer2012hyter} introduced an annotation tool and process
that can be used to create meaning-equivalent networks that encode an exponential 
number of translations for a given sentence.
Orthogonal to the
number of references, the quality of the reference translations is also essential to 
the reliability of automated evaluation~\citep{zbib2013crowd}. 
This topic itself raises the question of human translation assessment, which is beyond the scope of this paper~\citep{moorkens2018book}.

\textit{Meta-evaluation} studies the correlation between human assessments and automatic 
evaluations~\cite{callison-burch2006bleu,callison-burch2008meta-evaluation,callison-burch-2009-fast}.
Indeed, automatic evaluation is useful only if it rewards hypotheses perceived as fluent 
and adequate by a human. Interestingly, previous work \cite{bojar2016wmt} has shown that a 
higher correlation can be achieved when comparing similar systems than when comparing different 
types of systems, e.g.\ phrase-based vs neural vs rule-based. In particular, rule-based systems
can be penalized as they produce less common translations, even when such translations are fluent 
and adequate. Similarly, recent benchmark results comparing neural systems on high resource 
languages~\citep{bojar2018wmt,barrault2019wmt} have shown mismatches between the systems with highest
BLEU score and the systems faring the best in human evaluations. \citet{Freitag19,edunov19lmeval} study this mismatch
in the context of systems trained with back-translation~\citep{Sennrich16} and noisy back-translation~\citep{Edunov18}. They observe that systems
training with or without back-translation (BT) can reach a similar level of overlap (BLEU) with 
the reference, but hypotheses from BT systems are more fluent, both measured by humans and by a 
language model (LM). They suggest considering LM scores in addition to BLEU.

\citet{Freitag19,edunov19lmeval} point at {\it translationese} as a major source of mismatch 
between BLEU and human evaluation. Translationese refers to artifacts from the 
source language present in the translations, i.e. human translations are often less fluent than 
natural target sentences due to word order and lexical choices influenced by the source 
language~\citep{koppel2011translationese}. The impact of translationese on evaluation 
has recently received attention~\citep{Toral18,Zhang19,Graham19}.
In the present work, we are specifically concerned that the presence of translationese in the references 
might cause overlap-based metrics to reward hypotheses with translationese language more than 
hypotheses using more natural language. The question of bias to a specific reference has 
also been raised in the case of monolingual {\it human} 
evaluation~\citep{fomicheva-specia-2016-reference,ma-etal-2017-investigation}. The impact of 
translationese in test sets is related to but different from the impact of translationese in the 
training data~\cite{kurokawa09,Lembersky12adapting,bogoychev2019domain,riley2019translationese}.

In this work, we explore collecting a single reference translation, using human paraphrases to steer away as much as possible from biases in the reference translation  that affect the automatic metrics to prefer MT output with the same style (e.g.\ translationese).
Automatic methods to extract paraphrase n-grams~\citep{zhou2006paraphrasing} or full sentence paraphrases~\citep{kauchak-barzilay-2006-paraphrasing,bawden2020explicit,thompson2020automatic}
have been used to consider multiple references.
In contrast, we generate a single unbiased reference translation generated by humans instead of trying to cover a wider space of possible translations.
In contrast to human paraphrasing (our instructions asked for most diverse paraphrases), automatic paraphrasing are still far from perfect~\citep{aurko2019paraphrases} and mostly generate local changes that do not steer away from biases as e.g.\ introducing different sentence structures.

\section{Collecting High Quality and Diverse References}
\label{sec:refcollection}

We acquired two types of new reference translations: first, we asked a professional translation service to provide an additional reference translation. Second, we used the same service to paraphrase existing references, asking a different set of linguists.

\subsection{Additional Standard References}

We asked a professional translation service to create additional high quality references to measure the effect of different reference translations. The work was equally shared by 10 professional linguists. The use of CAT tools (dictionaries, translation memory, MT) was specifically disallowed, and the translation service employed a tool to disable copying from the source field and pasting anything into the target field. The translations were produced by experienced linguists who are native speakers in the target language.
The original WMT English$\to$German newstest2019 reference translations have been generated in sequence while keeping an 1-1 alignment between sentences. This should help the linguists to use some kind of document context. We instead shuffled the sentences to also get translations from different linguists within a document and avoid systematic biases within a document.
The collection of additional references not only may yield better references, but also allows us to conduct various types of multi-reference evaluation. In addition of applying multi-reference BLEU, it also allows us to select the most adequate option among the alternative references for each sentence, composing a higher quality set.

\begin{figure*}[htb]
  \begin{tabular}{|l|}
\hline
\underline{\bf{Task: Paraphrase the sentence as much as possible:}} \\
To paraphrase a source, you have to rewrite a sentence without changing the meaning of \\ 
the original sentence. \\

   \ \ \ \ \ \ \ \ 1. Read the sentence several times to fully understand the meaning \\
   \ \ \ \ \ \ \ \ 2. Note down key concepts \\
   \ \ \ \ \ \ \ \ 3. Write your version of the text without looking at the original \\
   \ \ \ \ \ \ \ \ 4. Compare your paraphrased text with the original and make minor adjustments to phrases \\
   \ \ \ \ \ \ \ \ \ \ \ \ \ that remain too similar \\
    
Please try to change as much as you can without changing the meaning of the original sentence. \\
Some suggestions: \\
\ \ \ \ \ \ \ \ 1. Start your first sentence at a different point from that of the original source (if possible) \\
\ \ \ \ \ \ \ \ 2. Use as many synonyms as possible \\
\ \ \ \ \ \ \ \ 3. Change the sentence structure (if possible) \\\hline
  \end{tabular}
  \caption{Instructions used to paraphrase an existing translation \emph{as much as possible}.}
  \label{fig:paraphrase_instructions}
\end{figure*}

\begin{table*}[th]
\begin{center}
{\setlength{\tabcolsep}{.2em}
\begin{tabular}{ ||c|l||}
\hline
 Source & \colorbox{yellow}{The Bells of St. Martin's} \colorbox{Apricot}{Fall Silent} \colorbox{Salmon}{as} \colorbox{CornflowerBlue}{Churches in Harlem} \colorbox{SpringGreen}{Struggle}. \\ \hline
Translation & \colorbox{Yellow}{Die Glocken von St. Martin} \colorbox{Apricot}{verstummen}, \colorbox{Salmon}{da} \colorbox{CornflowerBlue}{Kirchen in Harlem} \colorbox{SpringGreen}{Probleme haben}. \\ \hline
Paraphrase & \colorbox{SpringGreen}{Die Probleme} in \colorbox{CornflowerBlue}{Harlems Kirchen} \colorbox{Apricot}{lassen} \colorbox{yellow}{die Glocken von St. Martin} \colorbox{Apricot}{verstummen}. \\ \hline
Paraphrase & \colorbox{CornflowerBlue}{Die Kirchen in Harlem} \colorbox{SpringGreen}{k{\"a}mpfen mit Problemen}, \colorbox{Salmon}{und so} \colorbox{Apricot}{l{\"a}uten} \colorbox{yellow}{die Glocken von} \\
 & \colorbox{yellow}{St. Martin} \colorbox{Apricot}{nicht mehr}. \\ \hline
\end{tabular}
}
\end{center}
\vspace{-0.5em}
\caption{Reference examples of a typical translation and two different paraphrases of this translation. The paraphrases are not only very different from the source sentence (e.g.\ sentence structure), but also differ a lot when compared to each other.}
\label{table:paraphrase_example}
\vspace{-0.9em}
\end{table*}

\subsection{Diversified Paraphrased References}

The product of human translation is assumed to be ontologically different from natural texts \cite{koppel2011translationese} and is therefore often called translationese \cite{Gellerstam86}. Translationese includes the effects of interference, the process by which the
source language leaves distinct marks in the translation, e.g.\ word order, sentence structure (monotonic translation) or lexical choices. It also often brings simplification \cite{laviosa1997comparable}, as the translator might impoverish the message, the language, or both. 
The troubling implication is that a reference set of translationese sentences is biased to assign higher word overlap scores to MT outputs that produces a similar translationese style, and penalizes MT output with more natural targets \cite{Freitag19}. Collecting a different type of reference could uncover alternative high quality systems producing different styles of outputs.

We explore collecting diverse references using paraphrasing to steer away from translationese, with the ultimate goal of generating a \textit{natural-to-natural} test set, where neither the source sentences nor the reference sentences contain translationese artifacts. In an initial experiment on a sample of 100 sentences, we asked  linguists to paraphrase (translated) sentences. The paraphrased references had only minor changes and consequently only minor impact on the automatic metrics. Therefore, we changed the instructions and asked linguists to paraphrase the sentence \emph{as much as possible} while also suggesting using synonyms and different sentence structures. The paraphrase instructions are shown in Figure~\ref{fig:paraphrase_instructions}.
These instructions satisfy not only our goal to generate an unbiased sentence, but also have the side effect that two paraphrases of the same sentence are quite different. 
All our paraphrase experiments in this paper are done with these instructions.
One might be concerned that paraphrasing ``as much as possible" might yield excessive reformulation at the expense of adequacy in some cases. 
To compensate for this in the present paper, we collect adequacy ratings for all produced paraphrases. These ratings allow us to select the most adequate paraphrase from among available alternatives for the same sentence, which results in a composite high paraphrase set with strong adequacy ratings (see Table \ref{table:quality_rating}).
A paraphrase example is given in Table~\ref{table:paraphrase_example}. Even without speaking any German, one can easily see that the paraphrases have a different sentence structure than the source sentence, and  both paraphrases are quite different.

\section{Experimental Set-up}

\subsection{Data and Models}

We use the official submissions of the WMT 2019 English$\to$German news translation task~\citep{barrault2019wmt} to measure automatic scores for different kinds of references. We then report correlations with the WMT human ratings from the same evaluation campaign. We chose English$\to$German as this track had the most submissions and the outputs with the highest adequacy ratings. 

\subsection{Human Evaluation}
\label{sec:huma_eval}
We use the same direct assessment template as was used in the WMT 2019 evaluation campaign. Human raters are asked to assess a given translation by how adequately it expresses the meaning of the corresponding source sentence on an absolute 0-100 rating scale.
We acquire 3 ratings per sentence and take the average as the final sentence score. In contrast to WMT, we do not normalize the scores, and report the average absolute ratings.

\section{Experiments}
We generate three additional references for the WMT 2019 English$\to$German news translation task. In addition to acquiring an additional reference (AR), we also asked linguists to paraphrase the existing WMT reference and the AR reference (see Section~\ref{sec:refcollection} for details). We refer to these paraphrases as WMT.p and AR.p.

\subsection{Human Evaluation of References}
\label{subsec:combine}

It is often believed that the most accurate translations should also yield the highest correlation with humans ratings when used as reference for an automatic metric. For that reason, we run a human evaluation (Section~\ref{sec:huma_eval}) for all reference translations to test this hypothesis (Table~\ref{table:quality_rating}).
While all reference translations yield high scores, the paraphrased references are rated as slightly less accurate. We suspect that this may at least in part be an artifact of the rating methodology. Specifically, translations whose word order matches that of the source (i.e. translationese) are easier to rate than translations that use very different sentence structures and phrasing than the source sentence.
We generated our paraphrased reference translation with the instructions to modify the translations as much as possible. Therefore, the non-translationese, perhaps more natural, nature of the paraphrased translations make it more demanding to assign an accurate rating.

As a by-product of these ratings, we consider selecting the best rated references among alternatives for each sentence. Representing this method of combining reference sets with the HQ() function, we generate 3 new reference sets. These are (a) HQ(WMT, AR), abbreviated as HQ(R); (b)  HQ(WMT.p, AR.p), abbreviated as HQ(P); and (c) HQ(WMT, AR, AR.p, WMT.p), abbreviated as HQ(all 4). Interestingly, the combined paraphrased reference \emph{HQ(P)} has a higher human rating than WMT or AR alone.

\begin{table}[ht]
\begin{center}
\begin{tabular}{ ||l||c||}
\hline
 & adequacy rating \\ \hline \hline
 WMT & 85.3\\ \hline 
 WMT.p & 81.8 \\ \hline 
 AR & 86.7 \\ \hline
 AR.p & 80.8 \\ \hline \hline
 HQ(R) [WMT+AR] & 92.8 \\ \hline
 HQ(P) [WMT.p+AR.p] & 89.1 \\ \hline
 HQ(all 4) [all 4] & 95.3 \\ \hline
\end{tabular}
\end{center}
\vspace{-0.5em}
\caption{Human adequacy assessments for different kinds of references, 
over the full set of 1997 sentences.}
\label{table:quality_rating}
\vspace{-1.5em}
\end{table}

\subsection{Correlation with Human Judgement}

Table~\ref{table:correlation_wmt19_ende} provides the system-level rank-correlations (Spearman's $\rho$ and Kendall's $\tau$)\footnote{We used the scipy implementation in all our experiments: \url{https://docs.scipy.org/doc/scipy/reference/stats.html}} of BLEU (calculated with sacreBLEU \cite{post-2018-call}\footnote{BLEU+case.mixed+lang.en-de+numrefs.1+smooth.exp+test.wmt19+tok.intl+version.1.4.2}) evaluating translations of newstest2019 for different references. On the full set of 22 submissions, all 3 new references (AR, WMT.p, AR.p) show higher correlation with human judgment than the original WMT reference, with the paraphrased references WMT.p coming out on top.
Furthermore, each paraphrased reference set shows higher correlation when compared to the reference set that it was paraphrased from. 

\begin{table}[ht]
\begin{center}
\begin{tabular}{||l||l||c|c||}
\hline
Full Set (22)  & Reference & $\rho$  & $\tau$ \\ \hline \hline
  \multirow{4}{*}{single ref} & WMT & 0.88 & 0.72 \\ \cline{2-4}
 & AR & 0.89 & 0.76 \\ \cline{2-4}
 & WMT.p & \textbf{0.91} & \textbf{0.79} \\ \cline{2-4}
 & AR.p & 0.89 & 0.77 \\ \hline \hline
 \multirow{3}{*}{single ref} & HQ(R) & \textbf{0.91} & 0.78 \\ \cline{2-4}
 & HQ(P) & \textbf{0.91} & 0.78 \\ \cline{2-4}
 & HQ(all 4) & \textbf{0.91} & \textbf{0.79} \\ \hline \hline
 \multirow{3}{*}{multi ref} & AR+WMT & 0.90 & 0.75 \\ \cline{2-4}
 & AR.p+WMT.p & 0.90 & \textbf{0.79} \\ \cline{2-4}
 & all 4 & 0.90 & 0.75 \\ \hline
\end{tabular}
\end{center}
\vspace{-0.7em}
\caption{Spearman's $\rho$ and Kendall's $\tau$ for the WMT2019 English$\to$German official submissions with human ratings conducted by the WMT organizers.}
\label{table:correlation_wmt19_ende}
\vspace{-0.8em}
\end{table}

Although, the combined reference HQ(R) (Section~\ref{subsec:combine}) improves correlation when compared to the non-paraphrased reference sets (WMT and AR), not one of the three combined references HQ(R), HQ(P), HQ(all 4) shows higher correlation than the paraphrased reference set WMT.p. This result casts doubt on the belief that if references are rated as more adequate, it necessarily implies that such references will yield more reliable automated scores.

We further find that multi-reference BLEU (calculated with sacreBLEU) does not exhibit better correlation with human judgments either than single-reference BLEU or than the composed reference sets HQ(x). It is generally assumed that multi-reference BLEU yields higher correlation with human judgements due to the increased diversity in the reference translations. However, combining two translated reference sets that likely share the same systematic translationese biases does still prefers translationese translations. Interestingly, multi-reference BLEU with multiple paraphrases also does not show higher correlation than single-reference BLEU.
Combining {\it all 4} references with multi reference BLEU shows the same correlation numbers as the combination of \textit{AR+WMT}. As we will see later, the BLEU scores calculated with paraphrased references are much lower than those calculated with standard references. They have fewer n-gram matches, which are mostly only a subset of the n-gram matches of the standard references. Adding paraphrased references to a mix of standard references therefore has a small effect on the total number of n-gram matches, and as a consequence the scores are not much affected.

Note that the correlation numbers already appear relatively high for the full set of systems. This is because both Kendall's $\tau$ and Spearman's $\rho$ rank correlation operate over all possible pairs of systems. Since the submissions to WMT2019 covered a wide range of translation qualities, any metric able to distinguish the highest-scoring and lowest-scoring systems will already have a high correlation. Therefore, small numeric increases as demonstrated in Table \ref{table:correlation_wmt19_ende} can correspond to much larger improvements in the local ranking of systems.

\begin{figure}[ht]
\begin{center}
\includegraphics[width=0.49\textwidth]{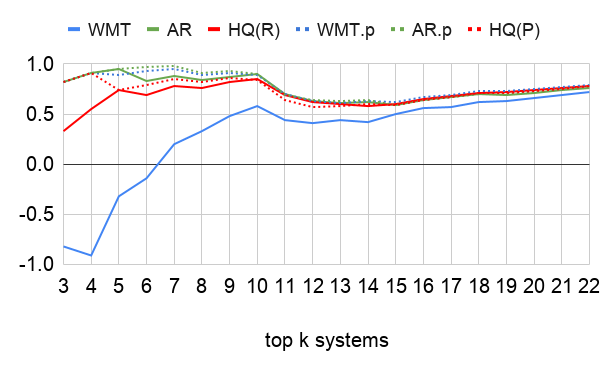}
\vspace{-2em}
\caption{Kendall's $\tau$ correlation of BLEU for the best k systems (based on human ratings). }
\label{fig:topk_systems}
\end{center}
\vspace{-1em}
\end{figure}

As a consequence, we looked deeper into the correlation between a subset of the systems that performed best in human evaluation, where correlation for metrics calculated on the standard reference is known to break down. 
Kendall's $\tau$ rank correlation as a function of the top k systems can be seen in Figure~\ref{fig:topk_systems}.
During the WMT 2019 Metric task~\citep{ma-etal-2019-results}, all official submissions (using the original WMT reference) had low correlation scores with human ratings. The paraphrased references improve especially on high quality system output, and every paraphrased reference set (dotted line) outperforms its corresponding unparaphrased set (same-color solid line).

These improvements in ranking can be seen in Table \ref{table:headsubmissions_wmt19_ende}, which reports the actual BLEU scores of the top seven submissions with four different references. Since we asked humans to paraphrase the WMT reference as much as possible (Section~\ref{sec:refcollection}) to get very different sentences, the paraphrased BLEU scores are much lower than what one expects for a high-quality system. Nevertheless, the system outputs are better ranked and show the highest correlation of any references explored in this paper.

\begin{table}[ht]
\begin{center}
{\setlength{\tabcolsep}{.2em}
\begin{tabular}{ ||l||c|c|c|c|c||}
\hline
 & WMT & HQ(R)  & WMT.p & HQ(P) & human \\ \hline \hline
FB & 43.6 & \textbf{42.3}& \textbf{15.1}& \textbf{15.0} & \textbf{0.347} \\ \hline
Micr.sd & 44.8 & 42.1 & 14.9 & 14.9 & 0.311 \\ \hline
Micr.dl & 44.8 & 42.2 & 14.9 & 14.9 & 0.296 \\ \hline
MSRA & \textbf{46.0} & 42.1 & 14.2 & 14.1 & 0.214 \\ \hline
UCAM & 44.1 & 40.4 & 14.2 & 14.2 & 0.213 \\ \hline
NEU	& 44.6 & 40.8 & 14.0 & 14.1 & 0.208 \\ \hline
MLLP & 42.4 & 38.3 & 13.3 & 13.4 & 0.189 \\ \hline
\end{tabular}
}
\end{center}
\vspace{-0.5em}
\caption{BLEU scores of the best submissions of WMT2019 English$\to$German.}
\label{table:headsubmissions_wmt19_ende}
\vspace{-1em}
\end{table}

\subsection{Alternative Metrics}

Any reference-based metric can be used with our new reference translations. In addition to BLEU, we consider TER~\cite{snover06ter}, METEOR~\cite{banerjee2005meteor}, chrF~\cite{popovic2015chrf}, the f-score variant of BERTScore~\cite{zhang19bertscore} and Yisi-1~\cite{lo2019yisi} (winning system of WMT 2019 English$\to$German metric task). Table~\ref{table:other_metrics} compares these metrics. As we saw in Figure \ref{fig:topk_systems}, the paraphrased version of each reference set yields higher correlation with human evaluation across all evaluated metrics than the corresponding original references, with the only exception of TER for HQ(P).
Comparing the two paraphrased references, we see that HQ(P) shows higher correlation for chrF and Yisi when compared to WMT.p. In particular Yisi (which is based on word embeddings) seems to benefit from the higher accuracy of the reference translation.

\begin{table}[ht]
\begin{center}
{\setlength{\tabcolsep}{.18em}
\begin{tabular}{ ||l||c|c|c|c|c||}
\hline
    metric & WMT &  HQ(R) & WMT.p & HQ(P) & HQ(all) \\ \hline \hline
    BLEU & 0.72 &  0.78 & \textbf{0.79} & \textbf{0.79} & \textbf{0.79} \\ \hline
    1\,-\,TER & 0.71 &  \textbf{0.74} & 0.71 & 0.67 & \textbf{0.74} \\ \hline
    chrF & 0.74 &  0.81 & 0.78 & \textbf{0.82} & 0.78 \\ \hline
    MET & 0.74 & \textbf{0.81} & \textbf{0.81} & \textbf{0.81} & 0.80 \\ \hline
    BERTS & 0.78 &  \textbf{0.82} & \textbf{0.82} & \textbf{0.82} & 0.81 \\ \hline
    Yisi-1 & 0.78  & 0.84 & 0.84 & \textbf{0.86} & 0.84\\ \hline
\end{tabular}
}
\end{center}
\vspace{-0.5em}
\caption{WMT 2019 English$\to$German: Correlations (Kendall's $\tau$) of alternative metrics: BLEU, 1.0 - TER, chrF, METEOR, BERTScore, and Yisi-1.}
\label{table:other_metrics}
\vspace{-0.5em}
\end{table}

\subsection{WMT18}
We acquired a paraphrased as-much-as-possible reference (WMT.p) for newstest2018 English$\to$German with the same instruction as used before (Figure~\ref{fig:paraphrase_instructions}).
The test set newstest2018 is a joint test set which means that half of the sentences have been originally
written in English and translated into German, and vice
versa. We paraphrased the reference sentences for the forward translated half only as we want to have a natural English source sentence. Correlation with human rankings of the WMT18 evaluation campaign are summarized in Table~\ref{table:wmt18}. The paraphrased reference WMT.p show higher correlations with human judgement for all metrics.

\begin{table}[ht]
\begin{center}
{\setlength{\tabcolsep}{.18em}
\begin{tabular}{ ||l||c|c|c|c|c||}
\hline
    ref & BLEU & chrf & METEOR & BERTS & Yisi-1 \\ \hline \hline
    WMT & 0.75 & 0.76 & 0.75 & 0.80 & 0.82 \\ \hline
    WMT.p & 0.91 & 0.82 & 0.84 & 0.90 & 0.91 \\ \hline
\end{tabular}
}
\end{center}
\vspace{-0.5em}
\caption{WMT 2018 English$\to$German: Kendall's $\tau$.}
\label{table:wmt18}
\vspace{-0.5em}
\end{table}

\section{Why Paraphrases?}

While the top WMT submissions use very similar approaches, there are some techniques in MT that are known to produce more natural (less translationese) output than others.
We run experiments with a variety of models that have been shown that their actual quality scores have low correlation with automatic metrics. In particular, we focus on back-translation~\cite{Sennrich16} and Automatic Post Editing (APE,~\citet{Freitag19}) augmented systems trained on WMT 2014 English$\to$German. All these systems have in common that they generate less translationese output, and thus BLEU with translationese references under-estimate their quality. The experiment in this section follows the setup described in \citet{Freitag19}.

We run adequacy evaluation on WMT newstest 2019 for the 3 systems, as described in Section~\ref{sec:huma_eval}. Both the APE and the BT models, which use additional target-side monolingual data, are rated higher by humans than the system relying only on bitext.  Table~\ref{table:bleu_ours} summarizes the BLEU scores for our different reference translations.
All references generated with human translations (WMT, HQ(R) and HQ(all 4)) show negative correlation with human ratings for these extreme cases and produce the wrong order. On the other hand, all references that rely purely on paraphrased references do produce the correct ranking of these three systems. 
This further suggests that reference translations based on human translations bias the metrics to generate higher scores for translationese outputs. By paraphrasing the reference translations, we undo this bias, and the metric can measure the true quality of the underlying systems with greater accuracy.

\begin{table}[ht]
\begin{center}
\begin{tabular}{||l||c|c|c|c||}
\hline
 Reference & bitext & APE & BT & correct? \\ \hline \hline
human & \cellcolor{lred1}84.5 & \cellcolor{lorange1}86.1 & \cellcolor{lgreen1}87.8 & \cmark \\ \hline
WMT & \cellcolor{lgreen1}39.4 & \cellcolor{lred1}34.6 & \cellcolor{lorange1}37.9 & \xmark \\ 
WMT.p & \cellcolor{lred1}12.5 & \cellcolor{lorange1}12.7 & \cellcolor{lgreen1}12.9 & \cmark\\ \hline
HQ(R) & \cellcolor{lgreen1}35.0 & \cellcolor{lred1}32.1 & \cellcolor{lorange1}34.9 & \xmark \\
HQ(p) & \cellcolor{lred1}12.4 & \cellcolor{lorange1}12.8 & \cellcolor{lgreen1}13.0 & \cmark\\
HQ(all 4) & \cellcolor{lorange1}27.2 & \cellcolor{lred1}25.8 & \cellcolor{lgreen1}27.5 & \xmark \\ \hline
\end{tabular}
\end{center}
\vspace{-0.6em}
\caption{BLEU scores for WMT newstest 2019 English$\to$German for MT systems trained on bitext, augmented with BT or using APE as text naturalizer. The {\it correct} column indicates if the model
ranking agrees with human judgments.}
\label{table:bleu_ours}
\vspace{-0.6em}
\end{table}

This finding, that existing reference translation methodology may systematically bias against modelling techniques known to improve human-judged quality, raises the question of whether previous research has incorrectly discarded approaches that actually improved the quality of MT. Releasing all reference translations gives the community a chance to revisit some of their decisions and measure quality differences for high quality systems.

\section{Characterizing Paraphrases}


\subsection{Alignment}
One typical characteristic of translationese is that humans prefer to translate a sentence phrase-by-phrase instead of coming up with a different sentence structure, resulting in `monotonic' translations. To measure the monotonicity of the different reference translations, we compute an alignment with
fast-align~\cite{dyer2013simple} on the WMT 2014 English-German parallel data and compare the alignments of all four references. Table~\ref{table:alignment} summarizes the average absolute distance of two alignment points for each reference. The paraphrased translations are less monotonic and use a different sentence structure than a pure human translation.

\begin{table}[ht]
\begin{center}
\begin{tabular}{||c|c|c|c||}
\hline
WMT & AR & WMT.p & AR.p \\ \hline \hline
5.17 & 5.27 & 6.43 & 6.88 \\ \hline
\end{tabular}
\end{center}
\vspace{-0.5em}
\caption{Average absolute distance per alignment point, as a proxy for word-by-word (`monotonic') translation. Lower scores indicate more monotonic translation.}
\label{table:alignment}
\vspace{-1.5em}
\end{table}

\subsection{Matched n-grams}

The actual BLEU scores calculated with the paraphrased references are much lower compared to BLEU scores calculated with standard references (Table~\ref{table:headsubmissions_wmt19_ende}). 
Nevertheless, the paraphrased references show higher correlation with human judgment, which motivates us to investigate which n-grams of the MT output are actually matching the paraphrased references during BLEU calculation.
The n-grams responsible for the most overlap with standard references are generic,
common n-grams. In the winning submission of the WMT 2019 English$\to$German evaluation campaign from Facebook, the 4grams with the highest number of matches are:
\begin{itemize}
    \setlength\itemsep{-0.2em}
    \item \textbf{, sagte er .} $\to$ 28 times (\textit{, he said.})
    \item \textbf{“ , sagte er} $\to$ 14 times (\textit{" , he said})
    \item \textbf{f{\"u}gte hinzu , dass} $\to$ 8 times (\textit{added that})
\end{itemize}
These matches are crucial to reach high $>40$ BLEU scores, and appear in translation 
when using the same sentence structure as the source sentence.  On the other hand, the n-grams overlapping with the paraphrased references show a different picture. They usually reward n-grams that express the semantic meaning of the sentence. The 4-grams with the highest number of matches with the paraphrased references for the same system are:
\begin{itemize}
    \setlength\itemsep{-0.2em}
    \item \textbf{Wheeling , West Virginia} $\to$ 3 times (\textit{Wheeling , West Virginia})
    \item \textbf{von Christine Blasey Ford} $\to$ 3 times (\textit{from Christine Blasey Ford})
    \item \textbf{Erdbeben der St{\"a}rke 7,5} $\to$ 3 times (\textit{7.5 magnitude earthquake})
\end{itemize}

\section{Conclusions}

This work presents a study on the impact of reference quality on the reliability of automated evaluation of machine translation. We consider collecting additional human translations as well as generating more diverse and natural references through paraphrasing. We observe that the paraphrased references result in more reliable automated evaluations, i.e. stronger correlation with human evaluation for the submissions of the WMT 2019 English$\to$German evaluation campaign. These findings are confirmed across a wide range of automated metrics, including BLEU, chrF, METEOR, BERTScore and Yisi. We further demonstrate that the paraphrased references correlate especially well for the top submissions of WMT, and additionally are able to correctly distinguish baselines from systems known to produce more natural output (those augmented with either BT or APE), whose quality tends to be underestimated by references with translationese artifacts.

We explore two different approaches to multi-reference evaluation: (a) standard multi-reference BLEU, and (b) selecting the best-rated references for each sentence. Contrary to conventional wisdom, we find that multi-reference BLEU does not exhibit better correlation with human judgments than single-reference BLEU. Combining two standard reference translations by selecting the best rated reference, on the other hand, did increase correlation for the standard reference translations. Nevertheless, the combined paraphrasing references are of higher quality for all techniques when compared to the standard reference counter part.

We suggest using a single paraphrased reference for more reliable automatic evaluation going forward. Although a combined paraphrased reference shows slightly higher correlation for embedding based metrics, it is over twice as expensive to construct such a reference set. 
%
To drive this point home, our experiments suggest that standard reference translations may systematically bias against modelling techniques known to improve human-judged quality, raising the question of whether previous research has incorrectly discarded approaches that actually improved the quality of MT. Releasing all reference translations gives the community a chance to revisit some of their decisions and measure quality differences for high quality systems and modelling techniques that produce more natural or fluent output.

As a closing note, we would like to emphasize that it is more difficult for a human rater to rate a paraphrased translation than a translationese sentence, because the latter may share a similar structure and lexical choice to the source. We suspect that human evaluation is also less reliable for complex translations. Future work, can investigate whether finer ratings could correct the bias in favor of lower effort ratings, and how this may interact with document-level evaluation.

\bibliography{emnlp2020}
\bibliographystyle{acl_natbib}

\end{document}